\documentclass[manuscript,screen]{acmart}

\AtBeginDocument{%
  \providecommand\BibTeX{{%
    \normalfont B\kern-0.5em{\scshape i\kern-0.25em b}\kern-0.8em\TeX}}}

\setcopyright{acmcopyright}
\copyrightyear{2018}
\acmYear{2018}
\acmDOI{XXXXXXX.XXXXXXX}

\usepackage{tikz}
\usetikzlibrary{mindmap}

\DeclareMathOperator*{\argmax}{argmax}
\DeclareMathOperator*{\argmin}{argmin}

\usepackage[OT2,T1]{fontenc}
\usepackage{substitutefont}

\substitutefont{OT2}{\rmdefault}{wncyr}

\DeclareRobustCommand{\rn}[1]{
	{\fontencoding{OT2}\selectfont#1}%
}

\begin{document}

\title{Modeling Human Behavior - Part I: Learning and Belief Approaches}


\author{Andrew Fuchs}
\affiliation{%
  \institution{Universit\'{a} di Pisa, Department of Computer Science}
  \city{Pisa}
  \country{Italy}}
\email{andrew.fuchs@phd.unipi.it}

\author{Andrea Passarella}
\author{Marco Conti}
\affiliation{%
  \institution{Institute for Informatics and Telematics (IIT), National Research Council (CNR)}
  \city{Pisa}
  \country{Italy}
}

\renewcommand{\shortauthors}{Fuchs, et al.}

\begin{abstract}
There is a clear desire to model and comprehend human behavior. Trends in research covering this topic show a clear assumption that many view human reasoning as the presupposed standard in artificial reasoning. As such, topics such as game theory, theory of mind, machine learning, etc. all integrate concepts which are assumed components of human reasoning. These serve as techniques to attempt to both replicate and understand the behaviors of humans. In addition, next generation autonomous and adaptive systems will largely include AI agents and humans working together as teams. To make this possible, autonomous agents will require the ability to embed practical models of human behavior, which allow them not only to replicate human models as a technique to ``learn", but to to understand the actions of users and anticipate their behavior, so as to truly operate in symbiosis with them. The main objective of this paper it to provide a succinct yet systematic review of the most important approaches in two areas dealing with quantitative models of human behaviors. Specifically, we focus on (i) techniques which learn a model or policy of behavior through exploration and feedback, such as Reinforcement Learning, and (ii) directly model mechanisms of human reasoning, such as beliefs and bias, without going necessarily learning via trial-and-error.
\end{abstract}



\maketitle






\section{Introduction}\label{sec:introduction}

Utilizing models of human behavior and decision-making has spanned decades and covered numerous approaches and applications. Human behavior and reasoning enables complex behaviors and social structures. Consequently, these structures become multifaceted and grow significantly in complexity. Still, humans are generally quite successful at navigating this complex social structure. There are a multitude of attempts at explaining aspects of these capabilities and this article discusses some of the more popular or persistent methods. The motivations behind research in the area of modeling human behavior and decisions are varied, so we limit the analysis to (a representative subset of) works providing \emph{quantitative} models (e.g. math models or algorithms), as these are the approaches that allow to ``code" human behavior in autonomous systems. Some examples include better autonomous driving \cite{wu2020joint, fernando2020deep}, comprehension of mental states \cite{bianco2019functional}, and population-level modeling \cite{jackson2017agent}. More generally, with the pervasive diffusion of AI systems and the recent focus on Human-Centric AI (HCAI, i.e., forms of AI where humans and AI agents work ``as a team"), embedding practical models of the human behavior that can guide the autonomous operations of AI systems will become a key required component of next-generation autonomous and adaptive systems.

A key distinction between the goals and approaches is often the fidelity of the replication and the expected deployment case. For instance, researchers may try to replicate the neurological pathways in an attempt to replicate the neuro-physical process underpinning reasoning \cite{asgaribrain}, or they may instead attempt to generate a computational model which is meant to mimic heuristically biased behavior \cite{lieder2017automatic}. In any case, a common aspect is the desire to use humans as the template for desirable patterns of reasoning. Using human reasoning and abilities has motivated numerous research topics enabling autonomous and adaptive systems. These systems can be trained to work independently, in a multi-agent system, and human-AI hybrid domains. In all of these cases, the resulting systems require the ability to both learn from the environment as well as be adaptive to changes observed. This enables the ability to act autonomously or gain and integrate new knowledge.

Regarding the need or desire to model human behavior, there are several topic areas which involve the interaction or dependence between humans and Artificial Intelligence (AI) systems. Humans are more frequently encountering intelligent systems and so it is vital that these systems be created with the use case and user in mind. In order to consider the user, it is important to understand their behavior and capabilities. As such, we will discuss methods for modeling and replicating human behavior. In this context, we will focus primarily on topics involving relevant concepts relating to Human-Centric AI such as AI-assisted decisions or Hybrid Intelligence. These topics demonstrate cases in which humans use or interact with AI systems. These interactions require differing assumptions regarding the dynamics between the human and the system and how those dynamics impact the capabilities and characteristics of the systems. In a more direct case of Human-AI Interaction (HAI) humans utilize the output of a system in their decision-making process. In this case, it is important the user clearly understands the output of an algorithm so they can effectively utilize the provided information. As an example, consider the use of AI or Machine Learning (ML) in medical diagnostics (i.e. Human-AI Interaction and AI-assisted Decision-Making). It is not only important that the system be effective and demonstrate high accuracy, but it must also prove useful for the human user. If not, the utility of the system may go unnoticed and unused.

In the most relevant case of HCAI for autonomous and adaptive systems, i.e., the case of hybrid intelligence, humans and AI systems form an interdependency. This can take multiple forms \cite{gurcan2021mapping, kaluarachchi2021review}, but leads to cases in which the human and the AI are expected to operate in a synergistic manner. Some examples include \cite{dellermann2021future}: Co-evolution over time, Human-in-the-loop learning, Interactive or Active Learning, Socio-technological ensemble. Given these paradigms, the human and AI can be paired in multiple configurations depending on technical, social, and other considerations. According to \cite{wilkensunderstandings}, there are five types of understanding associated with human-centered AI:
\begin{itemize}
	\item \textbf{Deficit-oriented:} AI serves to augment the human and compensate for deficiencies in attention, concentration, or physical and mental stamina.
	\item \textbf{Data reliability-oriented:} Provides AI as a tool for improving the use and understanding of data. (e.g. medical diagnosis)
	\item \textbf{Protection-oriented:} Use of AI to perform tasks too dangerous for the human or serves as an assistant to boost safety.
	\item \textbf{Potential-oriented:} Relates to combined intelligence to boost the performance and reasoning of the two sides as a team.
	\item \textbf{Political-oriented:} Relates to the concepts regarding how control, labor, etc. will be distributed between humans and technology, especially as it relates to protections for the employees.
\end{itemize}

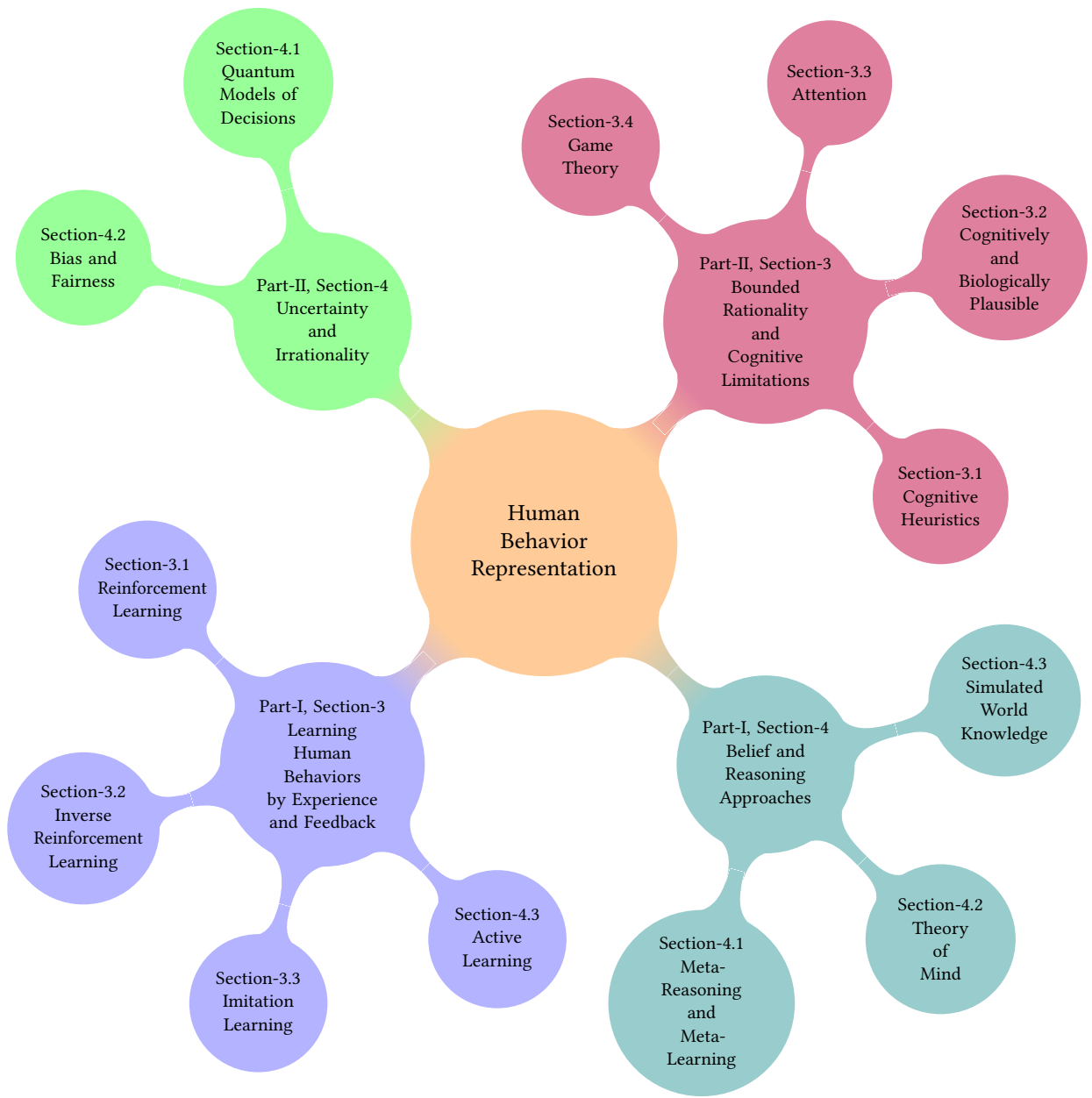
\begin{figure}
    \begin{tikzpicture}[mindmap, grow cyclic, text width=2.5cm, inner sep=1mm, every node/.style=concept, concept color=orange!40,
        level 1/.append style={level distance=4.75cm,sibling angle=90},
        level 2/.append style={level distance=3.75cm,sibling angle=60}]
    
    \node{Human\\Behavior\\Representation}
        child [concept color=blue!30, font=\fontsize{8pt}{10pt}\selectfont] { node {Part-I, Section-3\\Learning\\Human Behaviors\\by Experience\\and Feedback}
            child [font=\fontsize{8pt}{10pt}\selectfont]{ node {Section-3.1\\Reinforcement\\Learning}}
            child [font=\fontsize{8pt}{10pt}\selectfont]{ node {Section-3.2\\Inverse\\Reinforcement\\Learning}}
            child [font=\fontsize{8pt}{10pt}\selectfont]{ node {Section-3.3\\Imitation\\Learning}}
            child [font=\fontsize{8pt}{10pt}\selectfont]{ node {Section-4.3\\Active\\Learning}}
        }
        child [concept color=teal!40, font=\fontsize{8pt}{10pt}\selectfont] { node {Part-I, Section-4\\Belief and\\Reasoning\\Approaches}
            child [font=\fontsize{8pt}{10pt}\selectfont]{ node {Section-4.1\\Meta-Reasoning\\and\\Meta-Learning}}
            child [font=\fontsize{8pt}{10pt}\selectfont]{ node {Section-4.2\\Theory\\of\\Mind}}
            child [font=\fontsize{8pt}{10pt}\selectfont]{ node {Section-4.3\\Simulated\\World\\Knowledge}}
        }
        child [concept color=purple!50, font=\fontsize{8pt}{10pt}\selectfont] { node {Part-II, Section-3\\Bounded Rationality\\and\\Cognitive Limitations}
            child [font=\fontsize{8pt}{10pt}\selectfont]{ node {Section-3.1\\Cognitive\\Heuristics}}
            child [font=\fontsize{8pt}{10pt}\selectfont]{ node {Section-3.2\\Cognitively\\and\\Biologically\\Plausible}}
            child [font=\fontsize{8pt}{10pt}\selectfont]{ node {Section-3.3\\Attention}}
            child [font=\fontsize{8pt}{10pt}\selectfont]{ node {Section-3.4\\Game\\Theory}}
        }
        child [concept color=green!40, font=\fontsize{8pt}{10pt}\selectfont] { node {Part-II, Section-4\\Uncertainty\\and\\Irrationality}
            child [font=\fontsize{8pt}{10pt}\selectfont]{ node {Section-4.1\\Quantum\\Models of\\Decisions}}
            child [font=\fontsize{8pt}{10pt}\selectfont]{ node {Section-4.2\\Bias and\\Fairness}}
        };
    \end{tikzpicture}
    \caption{Taxonomy of concepts}
    \label{fig:taxonomy_of_concepts}
    \vspace{-5mm}
\end{figure}

The literature on modeling HCAI systems for achieving higher levels of autonomy and adaptation can be classified as show in Figure~\ref{fig:taxonomy_of_concepts}. Due to the vastness of contributions in the literature, we group work in two, consistent parts, where this paper (part I) covers the first two areas, i.e., approaches for automatically learning a behavior from examples, and approaches devoted to model belief and reasoning aspects. On the other hand, the remainder of the topics illustrated in Figure~\ref{fig:taxonomy_of_concepts} are covered in~\cite{fuchs2022partII}. Specifically, in this paper we will discuss some of the popular topics and applications relating to Human-Centric AI, Human-AI Interaction, and Hybrid Intelligence. These topics include: Reinforcement Learning, Meta-Learning, Theory of Mind, and more. These topics represent methods which attempt a model mimicking or inspired by differing biological/neurological, cognitive, and social levels of reasoning. Additionally, we will discuss how these topics align with application areas of interest. These application areas cover a wide assortment of both scenario as well as level of autonomy expected. For instance, this can include topics such as demographic preferences \cite{jackson2017agent} to something as safety-critical as fully autonomous driving \cite{wu2020joint, fernando2020deep}. The specific scenario can rely significantly on the level of autonomy expected and the level of risk or control humans are willing to allow. For these topics, we will provide underlying principles and definitions, relevant examples of use cases and their approaches, and further examples of relevant survey/review papers and related resources. Finally, we will provide additional details regarding some common application areas for these topics.

The rest of the paper will be structured as depicted in Figure~\ref{fig:taxonomy_of_concepts}. In Section~\ref{sec:learning_by_experience}, we will discuss learning methods which generate a model of behavior either by trial and error, or by learning from observations of others. These techniques learn from feedback denoting desirable behavior and adapt their policy according to this feedback. Next, in Section~\ref{sec:belief_reasoning_approaches}, we discuss methods which attempt to model the mental states of others, utilize world models to simulate human knowledge, or discuss bias and fairness in representations and reasoning. Similar to the previous section, these topics are focused on the bounded resources for reasoning and decision-making of humans, but in this case focus less on a direct replication of cognitive functions. Further, these techniques focus more on models of the world and those in them. Finally, Section~\ref{sec:conclusions} provides a brief overall comparison of the considered approaches.

We will not provide a comprehensive coverage of all aspects relating to the listed topics, but instead attempt to provide a useful assortment as a demonstration of topics of interest offering potential areas of further exploration. These topics and demonstrated applications are intended to illustrate situations in which humans interact with AI systems, how those interactions rely on an understanding or model of the human's behavior, and how systems can learn from humans or models of their abilities. In each of 
Sections~\ref{sec:learning_by_experience}-\ref{sec:belief_reasoning_approaches}, we discuss specific topic areas of interest (e.g. RL in Section~\ref{sec:learning_by_experience}),  
to provide examples of approaches and techniques illustrating aspects of these topics. Each section describing a 
topic is organized according to a common structure. First, we point to specific surveys and related dealing in greater detail with that topic. Then, we discuss the general principles. Next, we discuss one concrete example where those principles are made practical. Finally, we briefly mention additional examples where the same principles have been applied.

\section{HCAI orthogonal concepts and samples of Application Areas}\label{sec:applications_and_related}

In this section, we discuss some broad concepts related to HCAI, which cut across the various modeling methods presented in detail in the remainder of the paper. Moreover, we will briefly discuss popular application areas demonstrating uses of the techniques discussed in this paper. This list is not comprehensive, but serves to demonstrate topics which are likely more familiar and of immediate interest. The approaches used demonstrate methods which serve to replicate, model, or learn from human behavior and capabilities.


\subsection{Orthogonal HCAI concepts}
It is important to note that different issues and considerations arise from varying concerns. For instance, some examples include how it is important to ensure the AI system is not difficult to use or explain to users, difficult to manage or maintain, or perceived as creepy by the potential users (as noted above) \cite{yang2020re, eiband2021support}. Additionally, the systems will need awareness and capabilities supporting numerous types of intelligence such as social, emotional, physical, etc. in order to best understand and interact with humans \cite{cichocki2021future} while operating autonomously. Further, the reliability, correctness, and resulting impact of the system can be viewed as an important factor \cite{perrotta2020deep}.

A key factor in human-centric paradigms such as Human-AI Interaction or Hybrid Intelligence is the fact that the behavior of the human and the AI are assumed to impact each other \cite{rahwan2019machine}. In the case of hybrid intelligence, each side is providing a deeper aspect to the relationship. In general, we would require AI systems which can observe and understand humans in order to improve their behavior in this hybrid domain \cite{kambhampati2019challenges} via adaptability. As an example, in some cases of Active Learning (AL) or Reinforcement Learning (RL), a model is being learned with the help of, or in observation of, a human teacher \cite{puig2020watch, liu2019self, ramaraj2021unpacking, navidi2021new, najar2021reinforcement, holzinger2019interactive}. In such a case, we are expecting the AI system to continually refine both its model relative to the data samples while also improving its ability to know when to ask for help. Further, one could expect the human to use their understanding of the system implicitly or explicitly to improve the utility or informativeness of the samples as they observe the progression of the system \cite{schneider2020humans}. In some cases, the samples or information can come as human retellings of past experience \cite{kreminski2019evaluating}. This can also be reversed in the sense that the system can be designed to improve the types of responses in order to guide the human and improve the information received or queries of the user \cite{villareale2021understanding}.

Systems can also be designed in order to learn to work with the human as a team working in tandem. In such cases, it is often desired to have the system augment the abilities of the human or maintain autonomous control over an aspect of the task which would be more challenging for the human (and is less critical with respect to the larger goal). For instance, AI systems can be trained to assist the human in a navigation task \cite{reddy2018shared}. In such a case, it is observed that there are aspects of the problem which are much easier for either the human or the AI, so sharing the responsibilities allows for improved performance over either working independently. For more direct assistance to the human, in \cite{morrison2021social} we see an AI system being used to augment the senses of the user in order to assist visually-impaired users in a social context. This demonstrates a case in which the AI is much more closely integrated into the sensory systems of the human and is intended as an unobtrusive augmentation of their sensory capabilities. From another perspective, the goal could also be a system which can operate independently of the human as an additional agent in the environment \cite{wang2020too}. In these cases, the agent is expected to respond to the observed behavior of the human in order to assist or avoid interfering as both work to achieve their tasks. In either case, it is important to consider how the two (or more) are expected to interact and respond to observed behavior \cite{zahedi2021human, gao2021human}.

Another important aspect of hybrid intelligence would be the consideration needed for when and how to delegate control between the human or artificial/autonomous system, or determining which tasks can be performed without human intervention \cite{ning2021survey, raghu2019algorithmic}. There needs to be graceful handling in the event the human and AI system are both attempting to control the system. Additionally, there needs to be clear guidance regarding when either should be in control. For instance, it is important to understand the reliability of the system and where it is likely to encounter errors. Further, it is important to understand how the human perceives the error behavior of the system in order to anticipate its impact on the interaction \cite{bansal2019beyond}.

To support understanding and learning for the AI systems, it is also important to support methods for representing the observed behavior from the human in a manner which can be tractable and potentially simplify learning. As an example, \cite{xie2020learning} encodes the observations in a latent space and then learns a model of behavior corresponding to the latent representation. Such an encoding can allow the agent to abstract the observations to support connections between similar observations as well as other benefits. Additionally, systems require the means to observe and adapt to humans \cite{puig2020watch, schatzmann2006survey}. An important example is human-aware robot navigation \cite{moller2021survey, mavrogiannis2021core} and further human-robot interaction scenarios \cite{semeraro2021human}. In the navigation context, the motion, goals, and general behavior are crucial for the robot to successfully navigate the environment. Similarly, models can be generated to estimate or predict the feedback expected from a human collaborator or teacher in order to boost training of the AI system \cite{navidi2021new, navidi2020human}.


\subsection{Application areas}


\subsubsection{Robotics}

Robotics can utilize demonstrated human capabilities as well as behaviors to learn skills or behavioral policies. For instance, humans can provide demonstrations of behavior for a robot learner using a Policy-gradient RL policy, which can then be improved through practice and exploration by the learner \cite{akbulut2021acnmp}. Similarly, humans demonstrate the ability to adapt skills to variations and new scenarios. This adaptability has been explored in RL agents in robotics to enable generalization from initial demonstrations or exploration \cite{julian2020never, ramaraj2021unpacking}, which enables aspects of learning from demonstrations and an interactive learning process as seen in Imitation Learning, Inverse Reinforcement Learning, and Active Learning. Robot agents can also learn to anticipate the movement of humans or other artificial agents in order to compensate for their movements or rendezvous at a later position \cite{wang2020model, mavrogiannis2021core, moller2021survey}. This concept can also be extended to further topics in human-robot collaboration \cite{semeraro2021human}. Additionally, robot vision can be designed to replicate models of human vision mimicking foveation \cite{baron1994exploring}, which allows agents to observe visual stimuli with similar attention and focus to stimuli. These topics demonstrate how robots can be designed to learn from humans and also learn to operate in an environment alongside humans. The approaches for these solutions span multiple disciplines, including ones described in this paper (e.g. RL, Inverse Reinforcement Learning (IRL), etc.).

\subsubsection{Driver Prediction and Autonomous Driving}

There have been numerous examples of research performed to model and predict behavior in a driving scenario \cite{kolekar2021behavior}. These topics include methods which model pedestrian behavior \cite{choi2019drogon} or the behavior of other drivers \cite{fernando2020deep, bhattacharyya2020modeling, chandra2020stylepredict}. This type of modeling serves to train autonomous vehicles how to successfully drive while predicting or compensating for the behaviors of others through the use of models based on Inverse Reinforcement Learning, Imitation Learning, and related. This predictive power is essential so systems can anticipate and react to the non-uniform nature of human behavior. In addition to modeling the behavior of other drivers, systems have been investigated which can model the vehicle control behavior of drivers \cite{pentland1999modeling}. This allows for comprehension and modeling of driver control movements when operating a vehicle.

\subsubsection{AI in Games and Teaching}

In the area of video and serious games, AI is being considered with respect to multiple aspects. In one aspect, researchers and developers are investigating techniques to integrate AI into game development and game character behavior \cite{young2004architecture, xia2020recent, yannakakis2018artificial, bontchev2021personalization, zhao2020winning}. This allows for levels, players, etc. to be generated or controlled by adaptive behavior models to create a broader scope of experiences. These examples demonstrate uses of approaches such as planners (see Section~\ref{sec:belief_reasoning_approaches}), Reinforcement Learning, and related to improve the performance, adaptability, or creation of games. Non-Player Character (NPC) behavior can be supported by these techniques to generate policies based on historical gameplay data or learned models of play. This can be from learned models of behavior or based on past human player choices \cite{de2019simulating}. NPCs with these characteristics could prove better suited to respond to player choices and allow for more options. Similarly, the generation of levels and scenarios can also be expanded by allowing for more dynamic combinations of resources by the system.

\section{Learning Human Behaviors by Experience and Feedback}\label{sec:learning_by_experience}

In the following sections, we will discuss methods which learn patterns of behavior by exploration and observation of feedback. In Section~\ref{sec:reinforcement_learning}, we discuss learning agents which perform RL by observing rewards which denote the desirability of actions in a particular state or context. Extending this concept of feedback-based learning, we discuss IRL in Section~\ref{sec:inverse_reinforcement_learning}. These learning agents develop a model of behavior in an attempt to replicate the observed behavior of others. The agent attempts to estimate the feedback which generated the behavior observed and then learn a corresponding model of behavior. Similarly, in Imitation Learning (IL) (see Section~\ref{sec:imitation_learning}), agents also attempt to replicate observed behavior. In this case, the learner does not attempt to replicate the feedback, but instead attempts to directly learn a model of behavior matching the observations. The final section, Section~\ref{sec:active_learning}, discusses Active Learning, i.e., a learning paradigm in which the agent is able to query a teacher for feedback regarding a subset of input values. This allows the learner to learn based on an estimated confidence regarding different inputs and to utilize the expertise of the teacher to update their knowledge.

\subsection{Reinforcement Learning}\label{sec:reinforcement_learning}

\subsubsection{Relevant survey(s):}

For relevant survey papers and related, please refer to \cite{schatzmann2006survey}.

\subsubsection{Principles and Definitions}\label{subsec:RL.PrinciplesDefs}

RL is a method by which situations are mapped to actions so as to maximize a reward signal \cite{sutton2018reinforcement}. The maximization is performed by the learning agent through exploration of an environment and the possible actions. This exploration generates a feedback signal via rewards which the agent uses to learn the behaviors resulting in the most desirable feedback. The feedback received can be provided immediately or can be a result of a sequence of actions. For example, an agent could receive a reward for each step they make in an environment or simply a single reward at the end of a training session which corresponds to the outcome. The different parameters of the problem lead to numerous techniques for learning optimal behavior policies.

\paragraph{Markov Decision Process}

In RL, agents are attempting to solve a sequential decision process, which is represented by a Markov Decision Process (MDP). This representation allows for the components of a scenario to be formally modeled with underlying assumptions, such as dependence on past states. An MDP can be represented by the tuple $\{S,A,T,R,p(s_0),\gamma\}$ \cite{puterman2014markov}. $S$ refers to the states of the environment which can be traversed by executing actions from $A$. The execution of actions causes a transition between states, which follows the transition probabilities $T:S\times A\rightarrow p(S)$. As a means for feedback, the reward function $R:S\times A\rightarrow\mathcal{R}$ provides reward signals based on the selected action. Note, $R$ can also be defined with the inclusion of the resulting state $R:S\times A\times S\rightarrow\mathbb{R}$. Additionally, $p(s_0)$ defines the probabilities over initial states and $\gamma$ defines the discount parameter (defined in Section~\ref{subsec:rl_policy_learning}).

\subsubsection{Policies and Learning}\label{subsec:rl_policy_learning}

Given a scenario or MDP, an agent can be trained to find a policy $\pi:S\rightarrow p(A)$ which defines a likelihood of actions given current state. An agent's policy is learned through trial and error by exploring the given MDP and observing the rewards $r\in R$ provided as output. A policy is based on an estimate of action utility based on past observations and estimates of trajectories. The agent can generate an estimate of discounted return based on a discounted sum of future rewards:
\begin{equation}\label{eqn:rl_discounted_future}
    G_t=\sum_{k=0}^\infty \gamma^k R_{t+k+1}
\end{equation}
where $R_{t+k}$ denotes the reward observed time steps after time $t$ and $\gamma$ is the discount parameter. This describes a measure of estimated return based on observations of trajectories. The value of $\gamma$ determines how quickly the scale of future rewards decays, which impacts how strongly those observations impact the estimates. With this method of estimating returns, agents can generate a model of likely utility of actions in states. This concept is used to define a value function $\nu (s)$ given state $s$ where the assumption is that the agent starts in state $s$ and executes actions according to their policy $\pi$ in following states. The distinction in this case is the fact that the estimate is based on estimated behavior determined by a policy $\pi$. This means that the estimated value will consider the estimates of future values given the current state and expected trajectory of future states. In \cite{sutton2018reinforcement}, $\nu (s)$ is defined as:

\begin{equation}\label{eqn:rl_state_value_function}
    \nu_\pi(s) = \mathbb{E}_\pi\left[\sum_{k=0}^\infty\gamma^k R_{t+k+1}|S_t=s\right], \forall s\in S
\end{equation}

Similarly, the action value function is used to estimate the value of executing an action $a$ in a given state $s$:

\begin{equation}\label{eqn:rl_action_value_function}
    q_\pi (s,a) = \mathbb{E}_\pi\left[\sum_{k=0}^\infty\gamma^k R_{t+k+1}|s_t=s,a_t=a\right], \forall s\in S
\end{equation}

These are fundamental equations with respect to behavior policy learning via RL. Learning methods can be generated providing a means to learn a representation of value following Equation~\ref{eqn:rl_action_value_function}. This is done by utilizing the trial and error process of RL. Agents explore the environment and select actions (following a learning scheme - e.g. Epsilon Greedy) in order to observe the effect of actions. The effect is reflected in the rewards observed following an action or sequence of actions. These observed rewards are used in the estimated state-action value function in order to incrementally improve the estimated utility of actions. This is done by refining the estimate through a modeling process which uses immediate rewards and the current estimate of value to refine the estimate. This cycle is used to learn a policy through the feedback loop created by taking an action, observing an outcome, and updating the policy of actions accordingly. As an example, a temporal difference method can be used in Q-Learning to learn a state-action value function:
\begin{equation}\label{eqn:q_update}
    Q_{t+1}(s,a) = (1 - \alpha)Q_t(s,a) + \alpha[r + \gamma \max_{a'}Q_t(s',a')]
\end{equation}
where $\alpha$ is the learning rate used to discount the scale of the current estimate in the update to the estimated value. As can be seen, there is a recursive relationship between the current state's utility and the value of future states in the discounted $\gamma \max_{a'}Q_t(s',a')$ term. This enables the relationship between the value of the current state and the value of future states under the assumption of following the current policy.

\begin{figure}[ht]
    \centering
    \includegraphics[scale=.25]{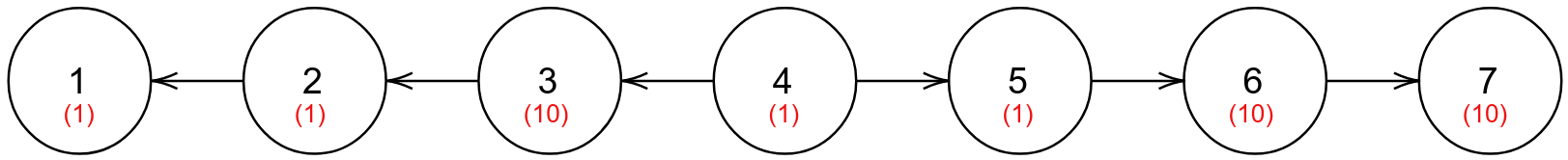}
	\caption{Sample Markov Decision Process showing issue with myopic view. State values are in parentheses below state labels.}
	\label{fig:myopicMDP}
\end{figure}

In RL, the method for defining and finding optimal behavior depends on the nature of the elements of the observation available to the agent. In the most general case, such as Q-Learning demonstrated above, the agent observes the current state and selects an action according to $\pi$. Typically, this is done by identifying $\argmax_a q_\pi (s,a)$. An important aspect of basing decisions on discounted future utilities is the possibility of MDPs which could result in sub-optimal behavior given a myopic view. For instance, Figure~\ref{fig:myopicMDP} provides an example of how too greedy a nearsighted view of state values would lead an agent to bias its behavior to lead from state $4$ to state $1$ with a sum of rewards equal to $12$ rather than the optimal choice, which is $4$ to $7$ with a sum of rewards equal to $21$. Such a scenario can be quite common and is one of the many challenges for successful policy learning in RL.

RL contains numerous examples of constraints which differentiate families of problems. For instance, it is possible to have an environment which is only partially observable \cite{shani2013survey}. In this case, not all aspects of a state are observable to an agent and so there is uncertainty regarding the exact state the agent is occupying. This results in a need for the agent to estimate the current state and use that estimation in the updates to the value function. In another case, the actions may not lead to an outcome with $100\%$ certainty. An example is a stochastic gridworld problem (e.g. OpanAI Gym FrozenLake \cite{1606.01540}). In this scenario, agents try to navigate a 2D world by moving up, down, left, or right. When an agent selects an action, there is a non-zero probability that the agent moves to an unintended state. For example, an ``up" move could in fact move the agent to the left. In this case, the learning method must support this uncertainty. It is worth noting that the previous examples represent scenarios supported by tabular methods. The use of RL has also been extended to continuous cases and there are methods utilizing Deep Learning. This is referred to as Deep RL and is commonly used in the case of continuous and/or large domains such as robotic control or video games \cite{arulkumaran2017deep}.

\subsubsection{Additional Relevant Results}

RL is an extensively studied and applied research topic. Agents have been trained to learn how to utilize attention or contextual information to improve policy learning \cite{salter2019attention, oroojlooy2020attendlight}. Agents can also be trained to utilize memories and past experiences in a more human-like manner in the hopes of solving credit assignment, handling sparse rewards, handling non-stationarity, comprehending relational knowledge, etc. \cite{mccallum1996hidden, hung2019optimizing, hung2019optimizing, lillicrap2019backpropagation, lansdell2019learning, padakandla2020reinforcement, wang2020instance, dvzeroski2001relational, forbes2002representations}. Similarly, RL agents can learn techniques to associate aspects of past experiences to contextual information for decision-making \cite{fortunato2019generalization, li2020cognitive}.

The use of past experience can also be extended to other agents or humans in the environment. An agent can learn to perform simultaneous control (i.e. shared autonomy) using RL methods \cite{reddy2018shared} or the observed behavior of others can be included in the observation space to allow for action selection conditioned on likely behavior of others \cite{xie2020learning}. An additional aspect of comprehending others is the prediction of behavior or goals, as is demonstrated in \cite{nguyen2020cognitive, nguyen2020effects}. Further, past experiences or behavior can be utilized as a means to predict likely future outcomes in order to guide agent behavior \cite{sun2019stochastic, zhang2020learning}.

\subsection{Inverse Reinforcement Learning}\label{sec:inverse_reinforcement_learning}

\subsubsection{Relevant survey(s):}

For relevant survey papers and related, please refer to \cite{arora2021survey, fernando2020deep}.

\subsubsection{Principles and Definitions}

IRL is a method by which an agent learns from examples of behavior without access to the underlying reward function motivating the behavior. The key distinction in this case is that the agent is trying to replicate or approximate the reward function $R_E$ or policy $\pi_E$ that caused the exemplar behavior. This results in effectively needing to learn a reward while simultaneously attempting to learn optimal behavior policy under the current estimated reward function. As such, the agent is performing two interdependent tasks. Given a policy $\pi_E$ or a set of $N$ demonstrated trajectories
\[\mathcal{D} = \{\langle (s_0, a_0),(s_1, a_1),\dots ,(s_j, a_j) \rangle_{i=1}^N: s_j\in S; a_j \in A; i, j, N \in \mathbb{N}\}\]
the agent is tasked with learning a representation which could explain the observed behavior \cite{arora2021survey}.

Generally speaking, there are numerous methods or approaches with respect to IRL. Therefore, we will be unable to address all the techniques in this section. We will instead provide some preliminary examples to provide some intuition regarding the common techniques and underlying principles. One method for the IRL task is that of apprenticeship learning \cite{abbeel2004apprenticeship}. In this case, there is an assumed vector of state-related features $\phi:S\rightarrow[0,1]^k$ which support the reward $R^*(s)=w^*\cdot\phi(s)$ with weight vector $w^*$. The feature vectors $\phi$ refer to observational data corresponding to the states (e.g. a collision detected flag). Given the definition of reward, the value of a policy $\pi$ can be measured by:
\begin{equation}\label{eqn:irl_policy_value}
    \mathbb{E}_{s_0\sim D}[V^\pi(s_0)] = \mathbb{E}[\sum_{t=0}^\infty\gamma^t R(s_t)|\pi] = w\cdot\mathbb{E}[\sum_{t=0}^\infty\gamma^t\phi(s_t)|\pi]
\end{equation}
with the initial states being drawn $s_0\sim D$ and with behavior following from the policy $\pi$. Then, the \emph{feature expectation} can be defined as:
\begin{equation}\label{eqn:irl_apprenticeship_feature_exp}
    \mu (\pi) = \mathbb{E}[\sum_{t=0}^\infty\gamma^t\phi(s_t)|\pi]
\end{equation}
which is used to define a policy's value $\mathbb{E}_{s_0\sim D}[V^\pi(s_0)] = w\cdot\mu(\pi)$. Given the estimation of feature expectation $\mu(\pi)$, the goal is to find a policy $\tilde{\pi}$ which can best match the observed demonstrations. In order to do so, this requires a comparison between $\tilde{\pi}$ and $\pi_E$. Since the policy $\pi_E$ is typically not provided, an estimate $\hat{\mu}_E$ based on demonstrations is needed. This can be accomplished by an empirical estimate:
\begin{equation}
    \hat{\mu}_E = \frac{1}{m}\sum_{i=1}^m\sum_{t=0}^\infty\gamma^t\phi(s^{(i)}_t)
\end{equation}
for a given set of trajectories $\{s^{(i)}_0, s^{(i)}_1,\dots\}^m_{i=1}$.

Given this structure, algorithms can be defined for iteratively refining the weight vectors and resulting policies. This enables the two components of the IRL paradigm. First, the estimated weight vectors $w^{(i)}$ define an estimated reward function which can be refined by updating the weight vector so as to reduce the discrepancy between the two estimated feature expectations $\hat{\mu}_E$ and $\mu(\pi)$ which measure agent performance. Second, the estimated reward with the current weight $w^{(i)}$ enables learning a policy which is optimal for the current estimate of policy value (Equation~\ref{eqn:irl_policy_value}) using RL, which is used in the first step to measure value discrepancy. This provides a cyclical process of reward refinement and policy learning.

It is important to note that the extra step of finding a suitable reward function and policy increases the complexity of the problem. The cycle of improving the reward requires the agent to retrain a behavior policy which reflects the new reward. On the other hand, finding an accurate representation of the reward would afford the agent a more general understanding of the behavior. Having access to a reward function allows an agent to understand desirable behavior at an abstract enough level to potentially transfer its understanding to a new environment. Of course, if the approximated reward isn't accurate enough, one would expect this to result in potential issues. In any case, there is the potential for increased generality of the resulting agent. It's worth mentioning that an exact replication of the underlying reward isn't entirely necessary. In fact, an affine transformation of the true reward function would in fact result in an equivalent policy \cite{arora2021survey}.

\subsubsection{Applications and Recent Results}

As an example of IRL, we refer to \cite{yang2020predicting} in which they try to learn a model to replicate human gaze patterns when searching for a visual target. The authors propose the use of imagery data with the human gaze fixations annotated. They used simulated fovea to learn a plausible model of desirable objects for attention. This is used to model how a human's gaze shifts around an image while attempting to locate an object within an image. The approach showed an improved ability to identify particular objects of interest rather than generating a saliency map to demonstrate attention.

In their method, they utilize an approach they refer to as Dynamic-Contextual-Belief (DCB), which is composed of three components: fovea, contextual beliefs, and dynamics. The fovea serves to mimic human vision and to provide only a sub-region in high detail while blurring or masking the remaining region. The masking results in a reduction in the observation space, limiting the input to a portion of the input space. In their approach, the masking is used to select a sub-region of the image to represent in high resolution while the remaining regions are represented using a blurred version of the image. This approximates the effect of the fovea on human vision and is used to represent the fixation of the observer. The contextual beliefs are used to represent a person's understanding of the contents of the scene such as objects and background items of an image. Lastly, the dynamics collects information regarding the focal fixations during search. The dynamics are represented as a transition between versions of the image which are updated based on iterations of fixation. Each region which receives fixation is replaced by its high-resolution representation resulting in a transition represented by:
\begin{equation}\label{eqn:irl_fovea_state}
    B_0 = L; B_{t+1} = M_t \odot H + (1 - M_t) \odot B_t
\end{equation}
where $B_t$ represents the belief state after $t$ fixations and $M_t$ is the mask generated by the $t^{th}$ fixation. $L$ and $H$ are belief maps which represent object and background locations for low-resolution and high-resolution images, respectively. Based on this definition, we can see that the representation of the image and the beliefs $B_t$  regarding contextual information and item locations are updated based on the iterative search conducted by moving the fixation around the image.

To represent the reward and learn a policy representing visual search behavior, the authors utilize Generative Adversarial Imitation Learning (GAIL). The GAIL framework utilizes the adversarial paradigm with networks representing the discriminator $D: S\times A \rightarrow (0,1)$ and generator $G$. These are used to train the system to generate data which matches the patterns of the original sampled data. The discriminator is tasked with learning to distinguish between real human data representing beliefs and fixation transition actions versus data which is generated artificially by $G$. The generator $G$ is tasked with learning to generate beliefs and actions which are convincing enough to the discriminator $D$ so as to be labeled as real rather than artificially generated. This is accomplished by maximizing an objective function:
\begin{equation}\label{eqn:irl_fovea_discriminator}
    \mathcal{L}_D = \mathbb{E}_r[\log(D(S,a))] + \mathbb{E}_f[\log(1-D(S,a))] - \gamma\mathbb{E}_r[||\nabla D(S,a)||^2]
\end{equation}
where $D(S,a)$ is the output of $D$ given the state-action pair $(S,a)$, $\mathbb{E}_r$ refers to the expectation over real state-action pairs, $\mathbb{E}_f$ refers to the expectation over fake search transition samples from $G$. The gradient term at the end of Equation~\ref{eqn:irl_fovea_discriminator} serves to improve the convergence rate. The definition of the reward is based on the output of the discriminator:
\begin{equation}\label{eqn:irl_fovea_disc_rew}
    r(S,a) = \log(D(S,a))
\end{equation}
For the generator, the performance uses Equation~\ref{eqn:irl_fovea_disc_rew} and is tasked with maximizing:
\begin{equation}
    \mathcal{L}_G = \mathbb{E}_f[\log(D(S,a))] = \mathbb{E}_f[r(S,a)]
\end{equation}
which shows that the higher likelihood of being real the discriminator places on generated data, the higher the resulting reward will be. To find an RL policy for the generator, the authors utilize Proximal Policy Optimization with the following representation:
\begin{equation}\label{eqn:irl_fovea_policy}
    \mathcal{L}_\pi = \mathbb{E}_\pi[\log(\pi(a|S))A(S,a)] + H(\pi)
\end{equation}
where the advantage function $A$ is estimated using generalized advantage estimation (GAE). The advantage represents the gain observed by taking action $a$ versus the policies default behavior. This definition of loss for the policy learning helps guide the learner toward actions which will result in higher advantage over other actions. The term $H$ is the max-entropy IRL term $H(\pi) = -\mathbb{E}_\pi [\log(\pi(a|S))]$ which helps improve convergence.

To test the approach, the authors utilized the COCO-Search18 dataset. The images were selected based on five criteria: Five criteria were imposed when selecting the TP images. First, they used no images portraying an animal or person so as to avoid the known strong biases to these categories. Second, they enforced each image having only a single instance of the target item. Third, the size of the target item must be within $[0.01, 0.1]$ the area of the image (measured by bounding box). Fourth, the target should lie outside the center of the image, which was determined by excluding items whose bounding box overlapped with a region in the center. Lastly, the image dimensions must have a ratio in $[1.2, 2.0]$ to allow for their display screen. Further filtering was performed and the reader can refer to the original text for more details.

The algorithm was tested against relevant baselines as well as against human performance. The authors utilized results from $10$ participants who viewed the $6,202$ images in the dataset. During their participation, eye tracking was performed during their search task. The algorithms compared against were: a random scanpath, a ConvNet detector, Fixation heuristics, a Behavior Cloning CNN, and a Behavior Cloning LSTM. The results demonstrated show their method meeting or exceeding the performance of the compared algorithmic methods. The results demonstrate the relationship between the number of fixation transitions made before finding the target and the success rate of the searches.

Additional tests were performed to test additional aspects, which can be found in the original text. As was demonstrated, this approach was able to successfully approximate the behavior demonstrated by the humans and also succeed at the identification tasks.

\paragraph{Additional Relevant Results}

As demonstrated, IRL provides an interesting approach to behavior modeling and replication. Agents can utilize observations of exemplar behavior in order to learn aspects of the desired model. For instance, agents can learn actions or skills from observed imagery via the inclusion of a learned cost function \cite{das2020model, wang2020learning} or by clustering observations into skills \cite{cockcroft2020learning}. Other recent results have focused on analytical/theoretical aspects of the problem relating to reward function search or aspects of the behavior policy providing the demonstrations \cite{balakrishnan2020efficient, kalweit2020deep, ni2020f, NEURIPS2020_19aa6c6f}. IRL can also be extended to methods which learn in a multi-task setting \cite{eysenbach2020rewriting} or observe policies of multiple other agents to operate and learn in a Multi-agent Reinforcement Learning (MARL) context \cite{reddy2012inverse, gruver2020multi}. Similar to the example in the previous section, IRL also proves useful in learning components of behavior leading to outcomes observed. For instance, learners can track or identify patterns which allow them to predict driver behavior \cite{wu2020joint}, simulate actions backward in time to predict likely trajectories leading to the current state \cite{lindner2021learning}, or predict gaze patterns for wheelchair drivers \cite{maekawa2020modeling}.

\subsection{Imitation Learning}\label{sec:imitation_learning}

\subsubsection{Relevant survey(s)}

For relevant survey papers and related, please refer to \cite{hussein2017imitation}.

\subsubsection{Principles and Definitions}

IL is a process which attempts to reproduce behavior given by experts in order to learn a pattern of behavior under a given task \cite{osa2018algorithmic}. At an abstract level, IL is closely related to IRL. In both cases, the goal is to utilize observations of behavior to train a behavioral model or policy. The key distinction comes in the structure of the learning process. In IRL, the learning process attempts to learn a suitable reward function which aligns with the demonstrated behavior. Additionally, the learning process uses the learned reward to generate a policy of behavior. The learned policy of behavior should closely model the demonstrated behavior. For IL, the learning process is not designed to generate a reward function which fits the demonstrations; instead, the process attempts to directly learn a model of behavior which best fits the demonstrated behavior.

As a demonstration of the principles, IL can be formulated as follows. Given a trajectory $\tau = [\phi_0,\dots,\phi_T]$ of features $\phi$, the learner is tasked with learning a policy reproducing the behavior. The vectors $\phi$ represent the features of the environment or system at each stage of the trajectory. The context, or state, $s$ representing the system state can be used in conjunction with an optional reward parameter $r$ as components of the underlying optimization problem.

With a set of demonstrations $\mathcal{D}=\{(\tau_i,s_i,r_i)\}_{i=1}^N$ of trajectories $\tau$, contexts $s$, and rewards $r$, (NOTE: the reward values $r_i$ might not be provided), the goal is to find a policy $\pi^*$. The policy should minimize the discrepancy between the distribution of features from the expert $q(\phi)$ and the distribution of features from the learner $p(\phi)$:
\begin{equation}\label{eqn:il_min_discrepancy}
    \pi^* = \argmin D(q(\phi),p(\phi))
\end{equation}
where $D$ is a discrepancy measure such as Kullback-Leibler. A key feature of Equation~\ref{eqn:il_min_discrepancy} is the fact that it promotes the alignment of behavior through direct observation by penalizing distributions with large discrepancies from the demonstration. This goal of low discrepancy guides the policy search toward those policies which best fit the observed distribution over features. Overall, the method for building will vary with each approach, but the underlying principle of direct replication is still a key component.

\subsubsection{Applications and Recent Results}

In \cite{liu2019self}, the authors demonstrate an IL method which is capable of performing policy improvement through the creation of examples without domain knowledge from a human. This allows the learner to perform policy improvement which may avoid the biases of the human domain knowledge commonly observed in IL tasks. To accomplish this, they define a policy improvement operator and methods for generating behavior exemplars. Note, this method differs from IRL as it is attempting to directly learn and refine a policy to fit the observed behavior without learning from an estimated reward function.

\paragraph{Preliminary Definition and Theory}

The following defines the general structure of the policy improvement operator $I$
\begin{equation}\label{eqn:il_generic_policy_improvement}
    \pi' = I(\pi,V_\pi)
\end{equation}
$I$ denotes an operator producing an improved policy $\pi'$ given a current policy $\pi$ and value function $V_\pi$. The improvement operator is described as a general black box operation, which means it will support multiple approaches. The key aspect from a theoretical standpoint is the constraint of policy improvement being satisfied, which means a newly generated policy must provide an improved estimated value. To denote this concept, they define the \emph{policy order} as $\pi \succ \pi'$ for policies $\pi, \pi'$ when $V_\pi(S) > V_{\pi'}$. This definition is used to prove the proposed approach will converge to an optimal policy given the operator by definition must output a policy with higher value (i.e. constraint of policy improvement). In other words, this is shown to be true in the event that the policy improvement operator satisfies $I(\pi, V_\pi) \succ \pi$.

With a policy defined using the policy improvement technique, the method can then provide improved demonstrations $\tau_{\pi'}$ for the IL process to create an updated policy $\pi'$. The learning method utilizes a loss function which measures the divergence between the current policy and the improved policy $L_I = D_{KL}[I(\pi,V_\pi)||\pi]$. In the case where the divergence is high, then the loss will be high and there is still room for an improved policy. Therefore, the utility is based on this measure of divergence to promote policy improvement. To estimate value, they implement a Deep Neural Network (DNN) for function approximation to provide an estimated value function given a state $s$: $\hat{V}(\theta, s) = DNN_\theta(s)$ (for an arbitrary $DNN_\theta$ - e.g. CNN, ResNet, etc.). They also define a policy estimator $\hat{\pi}(a|s,w)=DNN_w(s)$. With these components, they define their approach, self-improving Reinforcement Learning (SI-RL). Based on the above definitions and functions, define the loss function of the policy improvement operator as:
\begin{equation}\label{eqn:il_improvement_operator_loss}
    L_I(w,\theta,s) = D_{KL}[I(\hat{\pi}(\cdot|s,w),\hat{V}(\theta,s))||\hat{\pi}(\cdot|s,w)]
\end{equation}
with updates:
\begin{equation}
    w_{k+1} = w_k - I(\hat{\pi}(\cdot|s,w),\hat{V}(\theta,s))\nabla_w \log(\hat{\pi}(\cdot|s,w))
\end{equation}
\begin{equation}
    \theta_{k+1} = \theta_k I(\hat{\pi}(\cdot|s,w),\hat{V}(\theta,s))\nabla_\theta \log(\hat{\pi}(\cdot|s,w)) + \nabla_\theta ||\hat{V}(\theta,s) - R||^2
\end{equation}
for reward $R$.

\paragraph{Training via GAN}

As an alternative to the use of the KL-divergence, the authors propose integrating the use of a Generative Adversarial Network (GAN) into the training process for the imitation module. This integration is performed by defining a discriminator network $D$ to assign low loss to improved policies and high loss to the initial policies. The generator $G$ is then tasked with the generation of policies which will incur a low loss. This defines a method which allows the system to construct policies which improve performance while also defining a tool to guide the construction of policies by discouraging policies which do not create an improvement over past performance. This is formalized by:
\begin{equation}
    \min_w\max_D \mathbb{E}_{\pi_w}[\log D_\psi (s,a)] + \mathbb{E}_{\pi'}[\log(1 - D_\psi (s,a)]
\end{equation}
where $\pi_w$ is the trained policy, $\pi'$ is the improved policy, and $D_\psi (s,a)$ is the output of the discriminator network. The use of a GAN is also seen in the Section~\ref{sec:inverse_reinforcement_learning} example application, but the two differ in which components of the approach is modeled. In this case, the GAN is used specifically to learn how to generate better policies to meet the requirements of the improvement operator.

\paragraph{Results}

As noted in their proposed approach, the method defined supports a broad class of DNN-based systems for policy improvement. In their experiments, the authors test the use of trust region policy optimization (TRPO), Monte Carlo tree search (MCTS), and a cross entropy method (CEM). For test scenarios, the authors performed tests using the game Gomoku, miniRTS, and Atari games.

The game Gomoku is a zero-sum game and the authors test their MCTS-based implementation in $3050$ games. The agent is trained via self-play and tested it intervals against different levels of MCTS opponents. As is demonstrated, their agent is able to compete successfully against varying levels of opponents at the different training stages.

The next scenario tested, miniRTS, provides rule-based opponent for testing. The authors tested their SI-GARL and SI-RL approaches against the miniRTS opponent and compared performance to DQN, REINFORCE, and A3C implementations. The results demonstrate a strong performance against the miniRTS opponent. To further test their approach, the authors played their trained models against the traned models from the DQN, REINFORCE, and A3C implementations. These results again show strong performance.

\paragraph{Additional Relevant Results}

Similar to IRL, IL provides a useful means by which exemplar behavior can be modeled or replicated. Beyond the above example, additional scenarios include uses of IL for training of non-player characters in video games \cite{borovikov2019towards} or modeling driver behavior \cite{bhattacharyya2020modeling}. Similarly, this can be applied to observed behavior in video games \cite{ross2011reduction}. Additionally, as an example, multiple techniques have been investigated regarding the improvement or efficiency of agent training \cite{zolna2019task, chen2020bail, niu2020active, holzinger2019interactive}. This can include more efficient data sampling via AL-based or human-in-the-loop improvements \cite{niu2020active, holzinger2019interactive} or improved selection of exemplars via the estimate of utility \cite{chen2020bail}.

\subsection{Active Learning}\label{sec:active_learning}

\subsubsection{Relevant survey(s)}

For relevant survey papers and related, please refer to \cite{settles.tr09, ren2020survey, najar2021reinforcement, ramaraj2021unpacking}.

\subsubsection{Principles and Definitions}

AL is intended as an AI paradigm by which the learner is able to query an oracle regarding unlabeled data. The goal is to enable a more efficient usage of potentially costly data by allowing the learner to identify a representation of confidence or understanding in order to determine which items should be prioritized or require input from an oracle \cite{settles.tr09}. This is described as a characteristic similar to the concept curiosity in humans, since AL demonstrates a motivation to inspect items with less experience or certainty. This is related to how the system is able to decide what it wants to learn more about, not a true reproduction of human cognition. However, it has been argued that humans must regulate their priorities regarding curiosity in order to when and what to learn \cite{ten2021humans}. When prompted, the oracle can then provide labels for the queried samples, reducing the need to label larger sized datasets prior to the start of training.

There are numerous ways in which the learner can represent its understanding of the data to determine its confidence level. The representation is used to measure which would be the most desirable query. A possible definition could be the use of entropy to measure the confidence:
\begin{equation}\label{eqn:al_uncertainty_entropy}
    x^*_{ENT} = \argmax_x-\sum_i P(y_i|x;\theta)\log P(y_i|x;\theta)    
\end{equation}
for labels $y_i$, instance $x$, and model $\theta$. This provides an information-theoretic representation of the uncertainty. Specifically, the instance selected for the oracle feedback is the one corresponding to the largest entropy in the probability of assignment to the various labels (as per Equation~\ref{eqn:al_uncertainty_entropy}), i.e., the one for which the distribution of probabilities is closer to a uniform distribution, thus corresponding to the most uncertain assignment. Another approach is the instance with the \emph{least confident} labeling:
\begin{equation}\label{eqn:al_uncertainty_least_confident}
    x^*_{LC} = \argmin_x P(y^*|x;\theta)
\end{equation}
where $y^* = \argmax_y P(y|x;\theta)$ is the most likely class labeling. This promotes querying instances which have a best labeling with the least confidence.

The underlying principle in Equation~\ref{eqn:al_uncertainty_entropy} and Equation~\ref{eqn:al_uncertainty_least_confident} remains the representation of how well the model can relate instances to the labels based on a self-aware measure of certainty. The utilization of this notion of confidence is what allows for systems to exhibit behaviors akin to curiosity. This allows them to guide the learning process and reduce the reliance on large datasets.

\subsubsection{Applications and Recent Results}

As a demonstration of AL, we will discuss the approach outlined in \cite{navidi2020human, navidi2021new} in which the authors create a method of active imitation learning in a RL context. The proposed approach is a divergence from the above examples of approaches, but still utilizes the general principles of AL. In this case, the authors attempt to learn a model which can predict the oracle's responses for improved performance and to compensate for potential delays between agent action and oracle response. The prediction model enables the agent in its learning process and guides the policy learning. Unlike the form of AL described above, the oracle feedback instead provides labels of good or bad behavior, so the agent's policy learning phase will learn to avoid poorly performing actions rather than querying low-confidence items. This creates a method which combines concepts from IL, AL, and RL. This way, the behavior of the oracle encodes a model of the human's preferences with respect to agent behavior. Regarding the underlying algorithm, the authors propose a combination of SARSA and A3C with human-in-the-loop training.

As a means to provide feedback from the teacher, agents observe a binary signal indicating positive or negative view from the human teacher. This is in contrast to other methods which might provide more complicated or graded feedback by utilizing varying levels of good or bad (e.g. $[-100, -50, 50, 100]$). In order to compensate for variance in both reaction time and frequency of responses from the teacher (e.g. some might tire of giving feedback), the authors propose a feedback prediction system.

The feedback predictor or manager $FB$ is designed to learn and predict the feedback behavior of the human teacher. $FB$ is provided responses from $\{-1,1\}$ to signify satisfaction: $1$ or dissatisfaction: $-1$. The feedback prediction policy is denoted as:
\begin{equation}
    FB(O,A) = \psi^T \theta(O,A)
\end{equation}
where $FB(O,A)$ denotes the teacher feedback policy, $\psi$ are the policy parameters, and $\theta (O,A)$ represents a density function modeling the human response delay.

The feedback is then used to guide the training of the RL policies for the agent. The authors test their approach utilizing the SARSA and A3C algorithms. The feedback weights past observations to give a teacher-based representation of the outcomes. These methods are tested against in the Open-AI Gym environment and the authors illustrate results for the Cart-Pole and Mountain-Car environments. In both cases, we can see the proposed methods performing strongly in the environments and against the baseline methods.

\paragraph{Additional Relevant Results}

The concept of AL can also be extended to compensate for the use of multiple teachers with potentially heterogeneous behavior policies \cite{nguyen2020active}. In such a case, agents can encounter feedback from differing teachers. Consequently, the authors aim to learn models representing the policies of the different teachers to compensate for each pattern of behavior. Active learning can also be extended to concepts relating to heuristics when tasked to choose between options. As is demonstrated in \cite{parpart2017active}, agents can be trained via active learning to replicate human performance and replicate their tendencies regarding the decision model likely utilized by the humans.

\section{Belief and Reasoning Approaches}\label{sec:belief_reasoning_approaches}

In the following sections, we will discuss methods which use representations of belief or world knowledge to guide or assist in agent learning. Relating to the concept of cognitive frugality, we will discuss Meta-Reasoning and Meta-Learning in Section~\ref{sec:meta_reasoning_learning}. These concepts relate to how skills or knowledge are utilized based on the current context. Further, these concepts also relate to how and when cognitive resources should be applied in order to determine which system will select the course of action. Compensating for the behavior or preferences of others is another important aspect of modeling and replicating human behavior. The topic Theory of Mind (ToM), which we discuss in Section~\ref{sec:theory_of_mind}, attempts to generate a model of the mental states of others in order to anticipate and collaborate with others. Lastly, we discuss methods which integrate a model of knowledge or world dynamics in Section~\ref{sec:sims}. These methods attempt to replicate human understanding of the world in order to perform reasoning at a level which utilizes these models rather than attempting to learn them indirectly through interactions with the environment.

\subsection{Meta-Reasoning and Meta-Learning}\label{sec:meta_reasoning_learning}

\subsubsection{Relevant survey(s)}

For relevant survey papers and related, please refer to \cite{costantini2002meta, gershman2015computational, ackerman2017meta, griffiths2019doing, peterson1967man, khan2020literature, hospedales2020meta, wang2021meta}.

\subsubsection{Principles and Definitions}

Meta-reasoning and Meta-learning consider topics related to a level of abstraction which allows the algorithm to make determinations at two levels. In meta-reasoning, this is often referred to `reasoning about reasoning' \cite{costantini2002meta}. Similarly, meta-learning is often referred to as `learning to learn' \cite{khan2020literature}. These indicate the core aspect, which is the ability to perform introspection in order to dictate behavior. More concretely, this often relates to an ability to determine how resources or behavior policies will be allocated to perform a reasoning or learning task. This implies a notion of cost or effort regarding the task as well as an ability to identify the most suitable solution/behavior based on the context of the problem. The notion of allocation of resources for reasoning systems can also be related to examples in human cognition and the notion of bounded rationality \cite{zilberstein2011metareasoning}, which we discuss in Part II.

\paragraph{Meta-reasoning}

As noted in \cite{russell1991principles}, real agents are limited in capacity with respect to reasoning. Such a limitation manifests in both computational power as well as the time to decide and act. Further, the benefit or utility of an action can deteriorate over the time it takes to deliberate or to execute the action. Consequently, there is an implicit trade-off between the cost of an action (deliberation and execution) and its intrinsic utility. The ability to make judgments, consciously or otherwise, regarding the appropriate balance of these factors is a key aspect of meta-reasoning. In the context of computer-generated solutions, the common approach is to consider minimal time for an acceptable solution or maximizing outcome at the expense of time to completion. The ability to find an appropriate balance is one of the key motivations for investigating meta-reasoning topics.

From an algorithmic perspective, the meta-reasoning problem can be viewed as a method of optimizing the expected utility of performing an action versus the cost to do so. Following \cite{griffiths2019doing}, this can be expressed as
\begin{equation}\label{eqn:meta_reasoning_voc}
    \textrm{VOC}(c,b)=\mathbb{E}_{p(b'|b,c)}\left[\max_{a'}\mathbb{E}\left[U(a')|b'\right] - \max_a\mathbb{E}\left[U(a)|b\right]\right] - cost(c),
\end{equation}
where $b$ is the agent’s current belief, $b'$ is the refined belief resulting from executing computation $c$, and $\mathbb{E}\left[U(a)|b\right]$ is the expected utility of taking action $a$ under utility $U$ over the distribution of outcomes corresponding to belief $b$. Given the VOC, a rational agent should select the action which maximizes its value (or perform no action if $\textrm{VOC} < 0$). Unfortunately, calculating the VOC is costly, and so methods need to be developed to approximate it or define a similar measure of cost versus utility.

\paragraph{Meta-learning}

Similar to meta-reasoning, the case of meta-learning relies on an ability to perform introspection in order to understand how to best utilize the resources at hand. In this context, the resources are being applied to behavior learning and utilization. As noted in \cite{hospedales2020meta}, meta-learning involves improving a learning algorithm over multiple learning episodes at two levels. First, in base learning, a learning algorithm learns to solve a task based on the scenario parameters (e.g. image classification). For the second phase, meta-learning, an outer level algorithm improves the inner learning algorithm. The improvement is made so the resulting learned model improves one of the outer objectives. There can be multiple outer objectives resulting in the need for the algorithm to consider which model/algorithm to apply in a given context. Similar to meta-reasoning, the accrual of information or the time for deliberation can be costly. Consequently, the algorithm requires a method by which it can make a determination regarding the cost/benefit trade-off between the two \cite{gershman2015computational, griffiths2019doing}.

For a more formal characterization, we will refer to the conventions provided by \cite{vanschoren2019meta}. Consider the accrual of behaviors for given tasks $t_j\in T$, where $T$ is the set of all known tasks. Given a set of learning algorithms parameterized/configured by $\theta_i\in\Theta$ and evaluation measures $P_{i,j} = P (\theta_i, t_j)\in\mathbf{P}$, a meta-learner $L$ can be trained to predict recommended configurations $\Theta^*_{new}$ for a new task $t_{new}$. In this paradigm, it becomes a matter of learning a function $f:\Theta \times T \rightarrow \{\theta^*_k\}$, $k=1\dots K$ which can generate configurations $\theta^*_k$. This then allows for the creation of a \emph{portfolio} of configurations, which can be associated to tasks which they are best-suited. This paradigm allows for the generation and selection of configurations based on the experiences of the algorithm. As a result, the learner develops an understanding of what to learn and how to learn it.

\subsubsection{Applications and Recent Results}

In \cite{lange2020learning}, the authors investigate memory-based meta-learning (related to the concept of `learning to learn'). For their scenario, the key concept is an algorithm which can determine when to rely on a learning algorithm for generating behavior versus utilizing a heuristic. This enables agents develop policies when it is feasible given the time/computational constraints and rely on heuristics otherwise. Relating to human cognition and behavior, the authors note a similarity between the hidden activations of LSTM-based meta-learners and the recurrent activity of neurons in the prefrontal cortex. Generally speaking, allowing a split between fast/frugal mechanisms and slower/costly reasoning systems fits examples in human cognition and the notion of bounded rationality, and we discuss these related concepts in Part II.

On the topic of learning behavior policies, the authors note three features of particular interest regarding the dependence of the meta-learning algorithm on the meta-reinforcement learning problem:
\begin{itemize}
    \item Ecological uncertainty: How diverse is the range of tasks the agent could encounter?
    \item Task complexity: How long does it take to learn the optimal strategy for the task at hand?
    \item Expected lifetime: How much time can the agent spend on exploration and exploitation?
\end{itemize}
Based on their analysis, they showed non-adaptive behaviors are optimal in two cases: low variance across tasks in the ensemble, and when time constraints prevent sufficient time for exploration.

For the first scenario investigated, they test their approach on a two-arm Gaussian bandit task. It is noted that this allows for an analytical solution of optimality. The agent interacts with the environment by performing a series of $T$ arm pulls. The rewards for the two arms are represented by a deterministic reward of $0$ for the first arm and a Gaussian distribution with variable mean $\mu$ for the second. The mean is sampled $\mu\sim\mathcal{N}(-1,\sigma^2_p)$ (where $\sigma_p$ specifies the scale of ecological uncertainty) at the beginning of each episode, and then is used to define the reward function $r\sim\mathcal{N}(\mu,\sigma_l)$. It is noted that this variability controls how many arm pulls are needed to estimate the mean reward and $\sigma_l$ controls how quickly the agent can learn the policy since it controls the consistency of the observed rewards. The scale of $\sigma_p$ clearly determines how much uncertainty should be expected for the agent regarding the mean reward observed from the stochastic arm, which consequently scales the difficulty of estimating the expected utility of this arm. For $\sigma_l$, this determines how difficult it will be to estimate the mean based on observed rewards.

Given this definition, the optimal solution is determined analytically. The solution is based on the problem characteristics. Since the agent should perform $n$ pulls for exploration before exploiting its knowledge, the goal is to identify the optimal number of trials $n^*$ before concluding the exploration. The solution is as follows:
\begin{align}\label{eqn:meta_learning_optimal_pulls}
    n^* &= \argmax_n\mathbb{E}\left[\sum_{t=1}^T r_t|n,T,\sigma_l,\sigma_p\right]\\\nonumber 
    &= \argmax_n\left[-n + \mathbb{E}_{\mu,r}\left[(T-n)\times\mu\times p(\hat{\mu}>0)\right]\right],
\end{align}
where $\hat{\mu}$ is the estimated mean reward of the second arm after $n$ exploration trials.

Based on their experiments, the authors noted two distinct types of behavior. In the first, learning via exploration is the optimal behavior. In the second, the agent should instead not learn and exploit its knowledge. The value in stopping learning is attributed to two aspects: 1) small ecological uncertainty can make it very unlikely the first arm is better; 2) too high of variance causes the lifetime to be too short for valid learning. As a result, the authors note this leads to two consequences. First, the values of $\sigma_l$ and $\sigma_p$ create a threshold between learning and non learning of behaviors. This determines whether learning is suitable or if the agent should instead rely on exploitation of the existing knowledge. Second, the optimal strategy is dependent on the time allotted for learning. The amount of knowledge an agent can attain is strictly dependent on the variance parameters and the time they have to learn the dynamics of the arms. The results indicate the relationship between complexity and uncertainty with respect to the expected number of trials. There is a clear delineation between the two behaviors indicated in the results.

Given the above specification, the authors generated agents trained on the bandit scenario, which they subsequently compared to the theoretically optimal solution. The outcome indicates the meta-learning paradigm creates agents matching the analytical solution. Given this structure, they note that such an agent demonstrates the dual components which handle learning a behavior policy and utilizing a hard-coded choice which selects the deterministic arm (which is the optimal choice in expectation).

To test the proposed approach in a more general case, the authors then train and test an LSTM-based actor-critic agent in an ensemble of gridworld tasks. This scenario allowed them to investigate the impact of lifetime on the exploration strategies of the agents generated. Intuitively, in the case of a long lifetime, the agent has more opportunity to search for higher-valued goals where shorter lifetimes would necessitate greedier identification of goals. This was tested by placing goals of increasing value at farther distances from the start state.

As indicated by the results, the trained agent demonstrates a learned preference for goals based on the proximity and lifetime $T$ provided. The results indicate the agent is able to learn a priority for farther goals when there is more time to discover and return to the higher-valued goal. Similarly, the agent learns to focus on goals closer to the start state when there is a higher restriction on $T$ or when the agent should prioritize the goal states based on time remaining. The agent exploits its policy to find the highest value goal while there is time and then switches to the lower value state when there is only time for these path lengths. This indicates the agent is able to exploit different experiential knowledge while also identifying when to switch between exploration or learned behaviors. Based on the two scenarios, the authors demonstrate their method having the means to learn how to switch between learning and exploiting learned models. This demonstrates `learning to learn' and an ability to minimize the cost of behavior.

\paragraph{Additional Relevant Results}

As described above, a key aspect to meta-reasoning and meta-learning is the cost of actions and deliberation. In \cite{lieder2017learning, lieder2017strategy}, we can see this concept investigated in the learning process. Additional aspects such as the use of reinforcement learning, memory, and additional theoretical analysis can be seen in \cite{mikulik2020meta, xu2020task, oh2020discovering, zhen2020learning}.

\subsection{Theory of Mind}\label{sec:theory_of_mind}

\subsubsection{Relevant survey(s)}

For relevant survey papers and related, please refer to \cite{bianco2019functional, graziano2019attributing}

\subsubsection{Principles and Definitions}

ToM is what gives humans the capacity for reasoning about the mental states such as beliefs and desires for other agents in their environment \cite{baker2011bayesian, wang2021towards, rabinowitz2018machine, freire2019modeling}. In most, if not all, aspects of daily life, humans rely on and utilize their ability to estimate the mental state of others. One can imagine numerous scenarios they might encounter daily where they are able to interact with someone while relying on this type of reasoning. Something as simple as trying to anticipate on which side to pass another pedestrian requires you to utilize this technique. It is also easy to imagine a more complex scenario. For instance, poker players must reason about both the hand likelihoods as well as the mental state of their opponents to ensure a higher chance of success. In the case of poker, players can utilize methods of deception in an attempt to disguise their true mental state and gain an advantage. As a result, players would need to reason about the mental states of their opponents to guide their playing strategy.

ToM relies on multiple aspects of perception. A person will utilize the non-verbal cues, contextual clues, and other stimuli to form a picture of the world from the perspective of another person. ToM is what allows you to imagine yourself in someone else's place to model the likely next steps. Such a skill allows humans to perform better both as individuals and as part of a team. As with the walking and poker examples, one could also imagine numerous scenarios where people work together to accomplish a goal. People working together will of course utilize verbal and other forms of communication, but there is also a significant reliance on each member's ability to use a deeper understanding of the observed behaviors of their collaborators. Without this ability, we could expect that every aspect of these interactions would require explicit and comprehensive communication regarding all aspects of the interaction.

The representation of others is not necessarily exact, but can instead utilize an approximate and higher-level model \cite{rabinowitz2018machine}. Further, it can be argued that humans bias their models of others based on their own perspective. It is natural to expect a person's past experiences to impact their model of another person's perspective. Given a system to model the perspective of others, the natural extension is using this ability to generate recursive relationships between them \cite{wang2021towards}. For instance, the ability to model the mental state of others can be extended to model the other person's model of yourself. This means a person can generate a model of how they are perceived from another person's perspective. As is noted by \cite{freire2019modeling}, there are several notable approaches to address the problem of ToM. These include methods which rely on RL, Neural Networks, policy reconstruction, etc.

\subsubsection{Applications and Recent Results}

An aspect of ToM is the ability to predict the behavior of others in order to act accordingly. The authors of \cite{wang2020model} demonstrate a use case in which an RL-based learner predicts the movements of another agent to support rendezvous in a multi-agent environment. This is accomplished by generating a model for motion prediction, which supports a Hierarchical Predictive Planning (HPP) module.

Formally, the authors define the problem as a Decentralized Partially Observable Markov Decision Process (Dec-POMDP) $\mathcal{M}=\langle n,\mathcal{S},\mathcal{O},\mathcal{A}_{1,\dots,n},T,\mathcal{R},\gamma\rangle$ where $n$ is the number of agents. $\mathcal{S}$ denotes the set of states and $\mathcal{O}=[\mathcal{O}_1,\dots,\mathcal{O}_n]$ is the joint observation space, where $\mathcal{O}_i$ is agents $i$'s observations. The joint action space $\mathcal{A}_{1,\dots,n}$ is the combination of the agent action spaces $\mathcal{A}_i$ for agents $i\in\{1,\dots,n\}$, which define the actions available to the agents. The transitions are defined by the function $T:\mathcal{S}\times\mathcal{A}_{1,\dots,n}\times\mathcal{S}\rightarrow [0,1]$, which models the probability of transitioning between states given a joint action $a_{1,\dots,n}$. $\mathcal{R}$ is the reward function and maps states and actions to a reward $r\in\mathbb{R}$. As defined in RL, $\gamma$ is the discount factor for the learning process. The assumption in DecPOMDPs is that agents receive noisy observations of the environment and so the true state $s_i$ is unknown and is instead represented by the observation $O_i$. Consequently, the behavior relies on the estimated current state to identify a desirable action.

Agents are provided observations $\mathcal{O}_i=[\mathbf{p}_i,\mathbf{p}_{-i},\mathbf{o},\mathbf{g}]$ where $\mathbf{p}_i$ and $\mathbf{p}_{-i}$ refer to the agent positions, $\mathbf{o}$ are the sensor observations, and $\mathbf{g}$ is the agent's goal. The agents learn a predictive model of motion via a self-supervision algorithm. The systems are tasked with learning two models: \emph{self-prediction}$(\mathbf{f}_i)$ and \emph{other-prediction}$(\mathbf{f}_{-i})$. In this context, $\mathbf{f}_i$ and $\mathbf{f}_{-i}$ refer to self and other dynamics models respectively. Note that these models are the key point where ToM is used in this example as they enable modeling and prediction of others. These models are learned to generate a self-prediction of position $\Delta \mathbf{p}^{t+1}_i$ and observation $\Delta \mathbf{o}^{t+1}_i$, using a time window of past position which covers the previous $h$ steps:
\begin{equation}
    (\Delta \mathbf{p}^{t+1}_i,\Delta \mathbf{o}^{t+1}_i) = \mathbf{f}_i(\mathbf{p}^{t-h:t}_i,\mathbf{o}^{t-h:t}_i,\mathbf{g})
\end{equation}
Similarly, a model for other-prediction (i.e. other agent prediction) is defined as
\begin{equation}
    (\Delta \mathbf{p}^{t+1}_{-i},\Delta \mathbf{o}^{t+1}_i) = \mathbf{f}_{-i}(\mathbf{p}^{t-h:t}_{-i},\mathbf{o}^{t-h:t}_i,\mathbf{g})
\end{equation}
The authors note that the models do not depend on actions, but instead utilize observations of positions and posed conditioned on goals, which prevents needing to know the action space of others.

Given the predictive models $\mathbf{f}_i$ and $\mathbf{f}_{-i}$, a decentralized policy $\Pi_i$ is generated. This is done via the cross-entropy method (CEM) to convert goal evaluations into belief updates over potential rendezvous points. They note the intuition behind this approach is that each agent is simulating a centralized agent that fixes the goal of all agents, which are used to pre-condition the the motion predicted by $\mathbf{f}_i$ and $\mathbf{f}_{-i}$. These predictions are performed in a rollout of $T$ time steps into the future to predict the pose of agents. Based on the rollouts, the predicted goals are scored according to
\begin{equation}
    \mathcal{R}(\mathbf{p}_{1,\dots,n})=\begin{cases}
          0, \, &\, |\mathbf{p}_j-\mathbf{p_\mu}| < d, \, \forall j\in 1,\dots,n \\
	      \sum_{j,k\neq j} -|\mathbf{p}_k - \mathbf{p}_j|, \, & otherwise\\
     \end{cases}
\end{equation}
where $\mathbf{p_\mu} = \frac{1}{n}\sum_{k\in 1,\dots,n}\mathbf{p}_k$ and $d$ is a precision parameter. This emphasizes accuracy of predicted future states and smaller distances between agents at rendezvous points. Based on the predicted goal values, goal states are sampled via a normal distribution and the estimation process continues in order to favor goals which bring the agents closer together. This process generates the policy $\Pi_i$ which is used to complete the rendozvous without centralized control. With the above approach, the authors demonstrate a ToM method which utilizes the first level of ToM reasoning to predict likely agent behavior based on past observations and the current circumstances.

\paragraph{Results}

The authors demonstrate their algorithm's capabilities in simulated and real world environments. The simulated environments range in complexity starting with no obstacles and transition to environments with multiple obstacles. Similarly, the physical environments vary in obstacle complexity, type, and layout. The approach was tested against learned, decentralized, planning-based, and centralized baselines. For the learned baseline, the authors used the popular MADDPG algorithm. Planning was performed using the RRT planning system. For the centralized system, the midpoint, other agent's position, and random point are used. As is clear from the results, the proposed approach performs strongly compared against the baselines. This demonstrates the agents are able to perform the rendezvous without the need for centralized control. In the case of the real world environments, we can see a similar effectiveness of the proposed approach. The results indicate the approach is able to translate from simulated environments into the real world.

\paragraph{Additional Relevant Results}

Similar to the provided sample above, there are further examples in which it is desired to understand and predict behavior of other agents in a multi-agent setting. In \cite{shum2019theory} authors utilize composeable team hierarchies to generate policies in multi-agent settings based on the behavior patterns of others/groups. Similarly, \cite{kopf2020partner} demonstrates simulating the policies of others based on past experiences. Additionally, \cite{dissing2020implementing} demonstrates another use of ToM in the context of robotics. Further examples of behavior conditioned on the behavior or goals of others can be seen in \cite{morveli2021dealing, freire2019modeling, chandra2020stylepredict, oguntola2021deep}. For example, \cite{chandra2020stylepredict} demonstrates the use of ToM to predict driver behavior patterns. Additionally, \cite{lee2018answerer} demonstrates a question-answer system which tries to model the internal models to provide behavior which improves information gain based on estimated models.

\subsection{Simulating Human Knowledge of World for Learners}\label{sec:sims}

\subsubsection{Relevant survey(s)}

For relevant survey papers and related, please refer to \cite{ullman2020bayesian}

\subsubsection{Principles and Definitions}

The topics in this section do not represent a comprehensive list, but instead serve to illustrate a type of learning which relies on a level of world comprehension to accomplish the learning task. In this case, learning agents are provided models which allow them to understand a feature of the environment instead of requiring the agent to learn these features as well. For example, humans demonstrate a capacity for modeling the fundamental characteristics of an environment such as physics, compositional structure, etc. (e.g. \emph{Solidity} \cite{ullman2020bayesian}). With such an understanding, the components of a scenario can be considered when learning or utilizing a skill. Such a behavior is demonstrated in multiple aspects of daily life. Something as simple as understanding that gravity allows a person to pour water into a glass is something we take for granted. Comprehension of these aspects of the world enables a rich set of behavior. For artificial systems, such a comprehension is often not provided or demonstrated. To overcome this, researchers have investigated methods in which features such as physical, compositional, hierarchical, etc. are provided as a model for simulation or planning.

\paragraph{Physics Models}

Following the assumption that humans possess an internal model or understanding of aspects of physics (e.g. pushing an object over the edge of a table will cause it to fall), digital systems can be generated to replicate these behaviors for modeling and simulation. Engines such as Unity or MuJoCo \cite{todorov2012mujoco} demonstrate motion of objects in physical environments and can provide realism to object motion for observations made by artificial agents. These models of system behavior can provide an agent with an internal estimator for world dynamics to support prediction and planning \cite{allen2020rapid, ota2020towards, ota2021data}. This lets a learner generate predicted future states for reasoning and planning.

\paragraph{Planners}

Similar to a physics engine, the components and dynamics of a system can be modeled by describing the objects in the environment and how they can interact, be utilized, or modified. This allows for constraints to be specified for states which must be met in order for an action to be available for execution. The combination of the environment composition as well as the available actions allows for planning. Planning is utilized in order to convert the system state from the initial configuration to a goal state through a sequence of actions. A commonly used method for representation is via Planning Domain Definition Language (PDDL) \cite{aeronautiques1998pddl}. An environment specified in PDDL can then be analyzed by a solution system to identify a desirable sequence of actions for the specified goal. These solutions are generated to utilize the world model and the defined constraints. This enables identification of viable plans without the need for a learning process such as RL. On the other hand, policy generation via RL with planning-based trajectories can be performed \cite{zhi2020online}.

\subsubsection{Applications and Recent Results}

To demonstrate the use of a modeling system to enable deeper understanding, we will discuss \cite{zhi2020online}, which integrates PDDL into the model for planning based on different likely goals or trajectories. The inclusion of a planning system was intended as a method by which the system could account for sub-optimal or failed plans and incorporate them into estimates of future outcomes. Further, this approach was desired as a method which could account for the difficulty of the planning phase itself. The authors noted how many methods attempting to estimate goals fail to consider the difficulty of the planning portion of the process. Additionally, they note how most methods require the assumption of optimal behavior or Boltzmann-rational action noise. The assumption of optimality for all goals would therefore require computation of all goals in advance, which is intractable. In their approach, the authors use the integration of a planning system to represent a boundedly rational agent, where the bounds provide a resource limitation on planning and plan execution. This model provides a mechanism for Bayesian inference of plans/goals, even those with sub-optimal solutions, requiring backtracking, or irreversible failure. The limitation forces a constraint on the time or resources available to make a decision, which forces the agent at times to generate only a partial plan up to the level afforded by the constraints.

As noted above, the proposed approach represents the states, observations, and goals using PDDL and a variant supporting stochastic transitions named Probabilistic PDDL. This is accomplished by representing states and goals via predicate-based facts, relations, and numeric expressions in PDDL format. This allows for modeling of world state and actions, which are available when the provided preconditions are satisfied (e.g. current state, tool availability, etc.). The combination of predicates (e.g. relations such as `on' or `at') and fluents (e.g. fuel-level) allows the planning system to identify system state and available resources in order to generate a chain of actions and outcomes leading to a goal state. The actions result in a change in predicates and fluents, which signify the transition between world and system states. For a simple example, stacking blocks can be planned and considers the stacking agent's state as well as the current position of the blocks. A block which is covered by another would not be available for stacking and so the predicate's constraint would not be met for this block. On the other hand, uncovered or top-most blocks would be available, so one from this set would be available for selection for movement. With this representation, the observer has a prior over goals $P(g)$ specified via a probabilistic program over PDDL goal specifications. It is also noted that observation noise can be modeled for both the Boolean predicates and numeric fluents. This is accomplished by flipping predicate values or adding continuous noise with some probability (e.g. flipping a block's covered flag when covered by another block).

To represent bounded rationality, the authors define a budget
\begin{equation}
    \eta\sim\textrm{NEGATIVE-BINOMIAL}(r,q)
\end{equation}
where $r$ denotes the maximum failure count and $q$ denotes the continuation probability. Therefore, $\eta$ sets an upper bound on the solution search. If the bound is reached, the agent executes a partial plan leading to the most suitable state reachable from the found plans. This allows the agent to find the best plan they can given the limited resources available. The authors note that this model supports any planner capable of producing partial plans. For their scenario, the authors are operating in a gridworld environment and utilize a variant of the $A^*$ algorithm \cite{hart1968formal}. that makes search stochastic. They note the modification to the $A^*$ algorithm relates to the state successor component and modifies how best next states are weighted. If an agent reaches an unexpected state or the end of its plan, then a new plan is generated with a newly sampled budget $\eta$.

To test the proposed Sequential Inverse Plan Search (SIPS) approach, the authors utilize several environments with goal set $G$ and state space $S$:
\begin{itemize}
    \item \textbf{Taxi} ($|G| = 3$, $|S| = 125$) A taxi has to transport a passenger from one location to another in a gridworld
    \item \textbf{Doors, Keys, \& Gems} ($|G| = 3$, $|S| \sim 105$) An agent must navigate a maze with
doors, keys, and gems
    \item \textbf{Block Words} ($|G| = 5, |S| \sim 105$) Goals correspond to block towers that spell one of a set of five English words
    \item \textbf{Intrusion Detection} ($|G| = 20, |S| \sim 1030$) An agent might perform a variety of attacks on a set of servers (20 possible goals corresponding to a set of attacks on up to 10 servers)
\end{itemize}
They test their approach against a Bayesian IRL (BIRL) model generated by value iteration. This allows them to compare against an approach based on RL solutions.

As a demonstration of effectiveness, the performance of SIPS and BIRL were compared to human performance in goal prediction. The results indicate a strong match between the SIPS model and the demonstrated human performance. These results were generated by collecting human goal inferences on ten trajectories with six sub-optimal or failed with $N=8$ subjects. Human inferences were collected every six timesteps.

For evaluation of accuracy and speed, the authors tested with a dataset of optimal and non-optimal trajectories. The optimal trajectories were obtained via the $A^*$ algorithm and the non-optimal trajectories were generated using a replanning agent model with $r=2, q=0.95, \gamma=0.1$. They performed inference on these datasets with a uniform prior over goals. Based on their tests, they found good performance with $10$ particles per goal without the use of rejuvenation moves.

In both test cases, the proposed method demonstrates strong performance. Additionally, the performance versus computational cost is quite strong in the majority of cases in comparison with the baseline BIRL.

\paragraph{Additional Relevant Results}

Similar to the above example, the provision of a model or planner has been investigated in further settings. For example, \cite{pentecost2016using} demonstrates the combination of ACT-R and a physics engine to enhance the predictive power of the cognitive model to better replicate human behavior. Additionally, \cite{allen2020rapid, ota2020towards, ota2021data} demonstrate integrating a physics model into training of a learned behavior model. In \cite{ota2020towards, ota2021data}, this allows training of the behavior in a simulated environment followed by a transition to the real world. The agent can learn a approximate solution based on the simulator and then learn a translation which maps the simulation to the real world. In \cite{allen2020rapid}, the authors demonstrate a learner using a physics engine to learn how to manipulate tools and objects to achieve a sequence of actions to accomplish a task. This enables motion prediction in order to determine the best sequence based on likely trajectories.

\section{Conclusion}
\label{sec:conclusions}

In this paper, we have demonstrated methods focusing on multiple aspects of human behavior and cognition in addition to how humans interact with artificial systems. The types and end uses for these interactions vary, but a key concept is the ability to learn from or adapt to humans. The pervasive diffusion of AI systems provides exceptional opportunities to build next-generation autonomous and adaptive systems in several application areas (which we have briefly described in Section~\ref{sec:applications_and_related}). To gain full advantage of this opportunity, it is of paramount importance that AI systems ``learn to know and anticipate" the behavior of the involved users. This is a cornerstone of Human-Centric AI systems for autonomous and adaptive behaviors, which require to embed practical models of the human behavior that can be used either to interpret users' actions, and to anticipate their reactions with respect to (certain or likely) stimuli. Humans use multiple advanced skills daily to navigate the world and their interactions with others. Researchers have investigated several topics such as Theory of Mind, Inverse Reinforcement Learning, Active Learning, and more in an attempt to capture some of these capabilities and imbue systems with more advanced capabilities. This enables systems to better understand and operate in the world, including cases in which there are humans to account for and support, or when systems are expected to operate autonomously. Further, it enables systems to perform tasks more skillfully by replicating the level of skill demonstrated by humans. In addition to learning from humans, these techniques often serve to enable richer and more intuitive interactions between humans and artificial systems. This enables both the interactions as well as the system's ability to account for the human and potentially gain further knowledge as a result of the interaction. This approach gives systems a way in which to account for the cyclical nature of these relationships in order to improve performance and learn. 

In the case of Section~\ref{sec:learning_by_experience} (\emph{Learning Human Behaviors by Experience and Feedback}), the presented techniques serve as useful approaches for learning models of behavior when a learner has access to an environment and a method for receiving feedback. The feedback enables the learner to explore the environment and learn which behaviors prove most useful or desirable. These methods extend from discrete to continuous cases and varying levels of human demonstration or input. A possible downside to these approaches is that they often require large amounts of training samples as they tend to be sample inefficient. On the other hand, they are often demonstrated as viable techniques for training autonomous systems to operate and adapt to given environments. In the case of these learning methods (e.g. Section~\ref{sec:reinforcement_learning}-Section~\ref{sec:active_learning}), agents demonstrate the ability to translate experiences or demonstrated behavior into a performant policy of behavior. Often, these learned models are capable of meeting or exceeding the performance of the human counterpart. On the other hand, many of these techniques rely on exploration of the state-action space, which can be costly. Methods such as those discussed in Section~\ref{sec:active_learning} attempt to improve the efficiency of this exploration and the utility of the feedback or demonstrations utilized. In the case of Section~\ref{sec:inverse_reinforcement_learning} and Section~\ref{sec:imitation_learning}, the learner attempts to convert observations of demonstrated behavior into a policy to enable learning from examples. In general, the topics demonstrated illustrate a powerful technique for learning behaviors, but also demonstrate there are considerations necessary to ensure accurate outcomes and sufficient resources for training or learning.

Moreover, in Section~\ref{sec:belief_reasoning_approaches} (\emph{Belief and Reasoning Approaches}), we discuss approaches which represent methods reasoning or considering biases or beliefs of humans. In this context, the presented approaches learn or use models of belief. When considering the limitations on resources available for reasoning (e.g. Section~\ref{sec:meta_reasoning_learning}), agents can learn a belief regarding the current task and what behaviors are most suitable. This belief can guide the agent with respect to how and when to dedicate effort. For instance, this can be belief with respect to the mental state of others. This enables a model which can attempt to account for the unspoken aspects of human interaction, which can prove essential for autonomous systems operating in the same environment as human users or collaborators. These models, as seen in Section~\ref{sec:theory_of_mind}, demonstrate impressive reasoning levels. However, similar to what has been discussed above, these methods can require extensive training, which is costly. Further, these methods can often rely on the inclusion of domain knowledge or noisy observations, which can impact the level of performance or the ability to generalize to more realistic settings. We also discuss scenarios which allow learners to utilize world models to reason at a higher level, preventing the need to learn models of the world while simultaneously trying to learn to use the model. Similar to the previous topic, this also requires development of models, which can be costly and challenging. Also, the use of these models (as expected) requires the ability to generate such a model. The generation of such a model or modeling system can be highly effective, but again requires inclusion of domain knowledge or costly model generation. For instance, the use of a planning system/language as seen with PDDL (see Section~\ref{sec:sims}) typically relies on explicit definitions of resources, actions, goals, and constraints in order to allow the solution generation. This is not necessarily unreasonable, but it is a worthwhile consideration when determining the level of prior knowledge desired for your approach. As can be seen in many examples in literature, this inclusion of domain knowledge affords an effective solution for generating plans of action for a system to achieve a desired outcome.

Overall, we think that the approaches and goals presented motivate the topics discussed in this paper and serve to promote further investigation into how humans and artificial systems interact and learn from each other.

\section{Acknowledgments}\label{sec:acknowledgments}

This work was supported by the H2020 Humane-AI-Net project (grant \#952026) and by the CHIST-ERA grant CHIST-ERA-19-XAI-010, by MUR (grant No. not yet available), FWF (grant No. I 5205), EPSRC (grant No. EP/V055712/1), NCN (grant No. 2020/02/Y/ST6/00064), ETAg (grant No. SLTAT21096), BNSF (grant No. \rn{KP}-06-\rn{DOO}2/5).

\bibliographystyle{ACM-Reference-Format}
\bibliography{main}

\end{document}